%% file: sigir2023-chatGPT-eliciting-model-knowledge-vs-prompt-knowledge.tex
\documentclass[sigconf,natbib=true,anonymous=false]{acmart}
\usepackage{amsmath,nccmath,tabularx} 
\usepackage{algorithm}
\usepackage{algorithmic}
\usepackage{color}
\usepackage{graphicx}
\graphicspath{{images/}}
\usepackage{wrapfig,lipsum}
\usepackage{tabularx}
\usepackage{graphicx}
\usepackage{lscape}
\usepackage{subcaption}
\usepackage{latexsym}
\usepackage{pifont}
\usepackage{multirow}

\usepackage{enumitem}
\setlist{leftmargin=6mm}
\usepackage[T1]{fontenc}
\usepackage[utf8]{inputenc}
\usepackage[font=small,labelfont=bf]{caption}
\usepackage{subcaption}
\usepackage{booktabs}
\usepackage{listings}
\AtBeginDocument{%
  \providecommand\BibTeX{{%
    \normalfont B\kern-0.5em{\scshape i\kern-0.25em b}\kern-0.8em\TeX}}}

\copyrightyear{2023} 
\acmYear{2023} 

\setcopyright{acmlicensed}\acmConference[XXX '23]{Proceedings of the XXX}{XXX, 2023}{Somewhere}
\acmBooktitle{Proceedings of the XXX, 2023, Somewhere}

\acmPrice{15.00}
\acmDOI{xxxxxx}
\acmISBN{978-1-4503-8037-9/21/07}

\begin{document}

\title[How ChatGPT prompt knowledge impacts health answer correctness]{Dr ChatGPT, tell me what I want to hear: \\How prompt knowledge impacts health answer correctness.}



\author{Guido Zuccon}
\affiliation{%
	\institution{The University of Queensland}
	\city{Brisbane}
	\state{QLD}
	\country{Australia}}
\email{g.zuccon@uq.edu.au}

\author{Bevan Koopman}
\affiliation{%
	\institution{CSIRO}
	\city{Brisbane}
	\state{QLD}
	\country{Australia}}
\email{bevan.koopman@csiro.au}

\begin{abstract}
Generative pre-trained language models (GPLMs) like ChatGPT encode in the model's parameters knowledge the models observe during the pre-training phase. This knowledge is then used at inference to address the task specified by the user in their prompt. For example, for the question-answering task, the GPLMs leverage the knowledge and linguistic patterns learned at training to produce an answer to a user question. 	Aside from the knowledge encoded in the model itself, answers produced by GPLMs can also leverage knowledge provided in the prompts. For example, a GPLM can be integrated into a retrieve-then-generate paradigm where a search engine is used to retrieve documents relevant to the question; the content of the documents is then transferred to the GPLM via the prompt. In this paper we study the differences in answer correctness generated by ChatGPT when leveraging the model's knowledge alone vs. in combination with the prompt knowledge. We study this in the context of consumers seeking health advice from the model. Aside from measuring the effectiveness of ChatGPT in this context, we show that the knowledge passed in the prompt can overturn the knowledge encoded in the model and this is, in our experiments, to the detriment of answer correctness. 
This work has important implications for the development of more robust and transparent question-answering systems based on generative pre-trained language models.


\end{abstract}

\ccsdesc[500]{Information systems~Evaluation of retrieval results}
\keywords{Model selection, Dense retrievers}

\maketitle

\input{Sections/1_introduction.tex}

\input{Sections/2_related.tex}
\input{Sections/3_general_effectiveness.tex}
\input{Sections/4_evidence_biased.tex}

\input{Sections/6_conclusions.tex}

\bibliographystyle{ACM-Reference-Format}
\bibliography{bibliography}

\end{document}

%% file: Sections/1_introduction.tex
\section{Introduction}

Prompt-based generative language models, such as ChatGPT, can be used to answer complex natural language questions, often with impressive effectiveness. Measuring effectiveness is often done on standard question-answering datasets, directly using the question as the input prompt. Like a query in ad-hoc retrieval, the question is a proxy for a real human's information need. But one's information need is complex --- the question is but a very shallow insight into a much richer set of knowledge. 

With the advent of ChatGPT like models, user's are able to express far richer questions, divulging far more knowledge about their information need. This can be a double edged sword: on the one hand, far more information is available to the model to generate a good answer; on the other hand, incorrect, biased or misleading prompt information may derail the model's ability to give a good answer. While in some domain this may not matter, in this paper we investigate if this can have serious consequences in the health domain. In particular, we consider the case of a health consumer (i.e. a user with no or scarce health literacy) using ChatGPT to understand whether a treatment \textit{X} has a positive effect on condition \textit{Y}. 
Our evaluation is done using the TREC Misinformation track, a resource that contains complex health questions, often pertaining to misinformation regarding the efficacy of treatments. We consider two main experimental conditions: \texttt{(question-only)} ChatGPT is asked to provide an answer to a health question without further information provided to the model, \texttt{(evidence-biased)} ChatGPT is asked to provide an answer to a health question provided a prompt with information from a web search result (document) containing information about treatment  \textit{X} and condition \textit{Y}. We investigate cases where such document supports the use of the treatment (supporting evidence) and cases where the document dissuades from the use of the treatment (contrary evidence). We then compare the outputs of question-only and evidence-biased to understand the impact of the knowledge provided in the prompts with respect to the prior knowledge encoded in the model. Given these settings, we can then evaluate our two main research questions:
\begin{description}
  \item[RQ1 - General Effectiveness:] How effective is ChatGPT at answering complex health information questions?
  \item[RQ2 - Evidence Biased Effectiveness:] How does biasing ChatGPT by prompting with supporting and contrary evidence influence answer correctness?
\end{description}

The paper is divided in two parts around these two questions. For general effectiveness (RQ1), we show that ChatGPT is actually quite effective at answering health related questions (accuracy $80$\%). When biasing the prompt for evidence (with supporting or contrary evidence), we find that: (1) the prompt knowledge often is capable of overturning the model answer about a treatment, (2) when evidence contrary to the ground truth is provided and ChatGPT overturns an answer generated when no evidence is provided, this is often for the worse (overall accuracy $63$\%), i.e. incorrect evidence is able to deceive the model into providing an incorrect answer to a question that otherwise the model could answer correctly. 

Previous work has shown that engineering prompts has an important impact on the effectiveness of GPLMs like ChatGPT~\cite{liu2023pre}. This paper adds to that understanding by contributing that it is not just the ``form'' of the prompt that matters (e.g. the clarity of the instructions contained in the prompt), but also that the correctness of evidence contained in the prompt can highly influence the quality of the output of the GPLMs. This is  important when GPLMs are integrated in a retrieve-then-generate\footnote{Also called retrieve-then-read in the literature.} pipeline~\cite{chen2017reading,karpukhin2020dense}, where information related to the question is first identified from a corpus (e.g., the Web), and this is then passed to the model via the prompt to use to inform the model's output.




%% file: Sections/2_related.tex
\section{Related Work}
Our study aims to evaluate the impact on ChatGPT responses of knowledge provided in the prompt vs. the knowledge encoded in the model. The effect of prompting pre-training language models is attracting increasing attention, especially with respect to the so call practice of prompt engineering, i.e. finding an appropriate prompt that makes the language model solve a target downstream task.
Prompt engineering, as opposed to fine-tuning, doesn't modify the pre-trained model's weights when performing a downstream task~\cite{liu2023pre}. Prompt engineering is commonly used to enable language models, which have been pre-trained on large amounts of text, to execute few-shot or zero-shot learning tasks, reducing the need to fine-tune models and rely on supervised labels. In this prompt-learning approach, during inference, the input $x$ is altered using a template to create a textual prompt $x'$, which is then provided as input to the language model to generate the output string $y$. The typical prompt-learning setup involves constructing prompts with unfilled slots that the language model fills probabilistically to obtain a final string $\hat{x}$, which is then used to produce the output $y$~\cite{liu2023pre}. 

In our study we do not perform prompt-learning: we instead use the prompt to pass external knowledge to the model and measure how this changes the answers it generates. Related to this direction, but in the context of few-shot learning, \citet{zhao2021calibrate} observed that the use of prompts containing training examples for a GPLM (in particular GPT-3) provides unstable effectiveness, with the choice of prompt format, training examples and their order being major contributors to variability in effectiveness. We have similar observations in that in RQ2 we vary the document provided as evidence in the prompt and find that any two documents can have widely different effect on the answer of the model despite having the same stance about the topic of the question.

In this paper, we empirically study the impact of prompts in changing ChatGPT responses in the context of an health information seeking task. Other works have investigated the effectiveness of ChatGPT to answer health-related questions.

\citet{gilson2022well} presented a preliminary evaluation of ChatGPT on questions from the United States Medical Licensing Examination (USMLE) Step 1 and Step 2 exams. The prompt only included the exam question, and ChatGPT answer was evaluated in terms of accuracy of the answer, along with other aspects related to logical justification of the answer and presence of information internal and external to the question. The model performance were found to be comparable to a third year medical student.

\citet{nov2023putting} compared ChatGPT responses to those supplied by a healthcare provider in 10 typical patient-provider interactions. 392 laypeople were asked to determine if the response they received was generated by ChatGPT or a healthcare provider. The study found that it was difficult for participants to distinguish between ChatGPT and healthcare provider responses to patient inquiries, and that they were comfortable using chatbots to address less serious health concerns -- supporting the realistic setting of our study.

\citet{benoit2023chatgpt} evaluated the correctness of ChatGPT in the diagnosis and triage of medical cases presented as vignettes\footnote{A clinical vignette is a brief patient summary that includes relevant history, physical exam results, investigative data, and treatment.} and found that ChatGPT displays a diagnostic accuracy of 75.6\% and a triage accuracy of 57.8\%. 

\citet{de2023chatgpt} recognises that ChatGPT can be used to spread misinformation within public health topics. In our study, we show how misinformation can be injected into ChatGPT's input through the prompt and this sensibly impacts the output of the model, with potentially dire consequences. 

%% file: Sections/3_general_effectiveness.tex
\section{RQ1 - General Effectiveness}
Our first research question relates to how effective ChatGPT is in answering complex health information questions. Measuring this serves two purposes: on one hand it informs about the reliability of this GPLM in supporting information seeking tasks related to health advice, on the other hand it allows us to ground the effectiveness of the model when relying solely on the knowledge encoded in the model itself, i.e. without observing any additional knowledge in the prompt (i.e. the prompt only contains the question).

\subsection{Methods}

We consider a total of 100 topics from the TREC 2021 and 2022 Health Misinformation track. Each topic relates to the efficacy of a treatment for a specific health issue. The topic includes a natural language question; e.g., “Does apple cider vinegar work to treat ear infections?” and a ground truth answer, either `helpful` or `unhelpful` for 2021 and  ‘yes’ or ‘no’ for 2022. Ground truth answers were assigned based on current medical practice.

We issue the question to ChatGPT as part of the prompt shown in Figure~\ref{fig:rq1_prompt}, which instructs the model to provide a Yes/No answer and associated explanation. We then evaluate the correctness of the answer by comparing ChatGPT answer to the TREC Misinformation Track ground truth. 

\begin{figure}
	\begin{verbatim}
	[question_text] 
	Answer <Yes> or <No>, and provide an 
	explanation afterwards.	
	\end{verbatim}	
	\caption{GPTChat prompt format for determining general effectiveness (RQ1) on TREC Misinformation topics.}
	\label{fig:rq1_prompt}
\end{figure}

\subsection{Results}

\begin{figure}     
     \begin{subfigure}{0.66\columnwidth}
         \includegraphics[width=1\columnwidth]{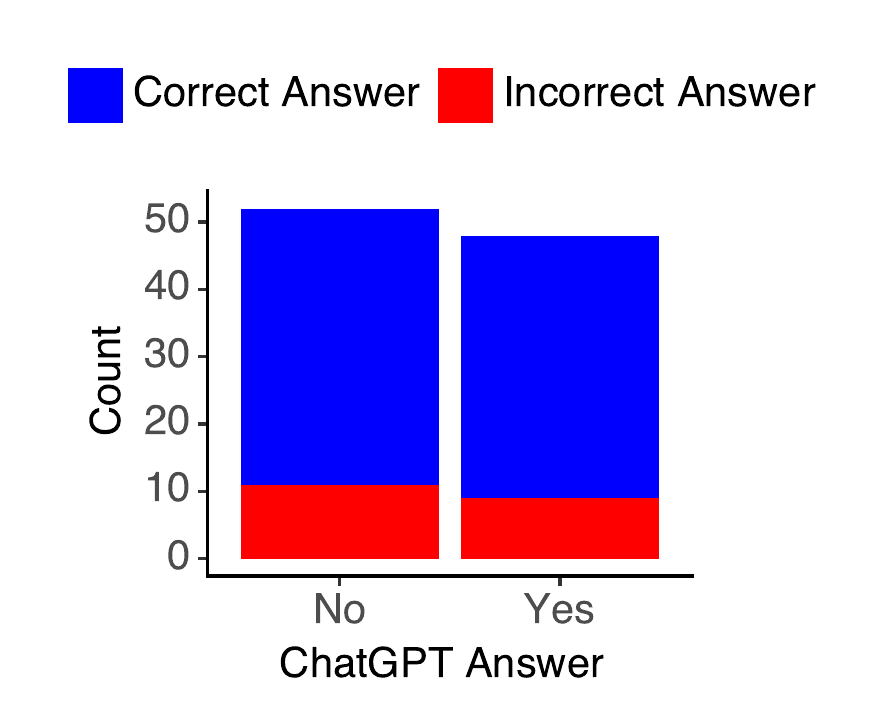}
     \end{subfigure}
     \begin{subfigure}{0.3\columnwidth}
         \centering
			\begin{tabular}{lr}
			\toprule
			  Correct\#   &    80   \\
			 Incorrect\# &    20   \\
			 Total\#     &   100   \\
			 Accuracy  &     0.8 \\
			\bottomrule
			\end{tabular}	         
      \end{subfigure}
  \caption{Effectiveness of ChatGPT when prompting for "Yes/No" answers to TREC Misinformation questions. }
  \label{fig:rq1_results}
\end{figure}

The effectiveness of ChatGPT is shown in Figure~\ref{fig:rq1_results}. Overall effectiveness was 80\%. ChatGPT answered "Yes" and "No" a similar number of times and its error rate was similar when answering "Yes" and "No". With this simple prompt, ChatGPT always provided an answer of Yes or No -- as opposed to when using the prompt of RQ2, as we shall discuss below. When generating an answer, ChatGPT also produced an explanation to support its stance. We did not perform an in depth analysis of these explanations, e.g., to verify whether the claims made in the explanation are true or are hallucinations~\cite{ji2022survey,bang2023multitask}, because we lack the medical expertise necessary for this. We plan to investigate this aspect in future work through engagement with medical experts. A brief analysis of these explanations, however, reveals that they often contain remarks about the presence of limited scientific evidence (or even conflicting evidence) with respect to a treatment option for a condition, and details about specific conditions for which the answer was not valid. The answers also often contain a suggestion to contact a health practitioner for further review of the advice.

%% file: Sections/4_evidence_biased.tex
\section{RQ2 - Evidence Biased Effectiveness}

Our second research question relates to the impact on answer correctness when biasing ChatGPT by prompting with (1) supporting evidence; and (2) contrary evidence. Measuring this allows us to determine the impact of the prompt knowledge and whether this can overturn the knowledge encoded in the model when generating an answer.

\subsection{Methods}

Supporting and contrary evidence is taken as individual document qrels in the TREC Misinformation track. Documents judged at $2$ (``Supportive``) are selected as supporting evidence. Documents judged as $0$ (``Dissuades``) are taken as contrary. The process to issue the evidence biased prompt to ChatGPT is as follows:
\begin{enumerate}
  \item For each TREC topic, we select a maximum of 3 supportive evidence and 3 contrary evidence documents from the TREC qrels;\footnote{Some topics did not contain 3 contrary or supportive documents: in these cases we selected as many as there were available.}
  \item For each document, we generate the prompt shown in Figure~\ref{fig:rq2_prompt}, including both the question and the evidence text in quotes;
  \item We capture the response from ChatGPT;
  \item Although the prompt explicitly asks for a Yes/No answer, we found that in many cases ChatGPT did not use these words, but did provide the answer in another way (e.g., ``Inhaling steam can help alleviate the symptoms of a common cold''). Thus we had to manually read the responses and determine if the answer was Yes, No, or Unsure. All documents were assessed by two people and discussion between the annotators took place to resolve label disagreement.
  \item Once ChatGPT provided an answer, we then evaluate the correctness of the answer by comparing ChatGPT's answer to the TREC Misinformation Track ground truth. 
\end{enumerate}

We used the 35 topics from TREC 2021 Health Misinformation track that contained document-level relevance assessments. (Document-level qrels are not available for TREC 2022 at the time of writing.) 

The actual documents are web pages from the noclean version of the C4 dataset: $\approx$1B English text extracts from the April 2019 snapshot of Common Crawl. Some  passages are long and exceed the maximum token limit of ChatGPT. We trimmed long texts using the NLTK \texttt{word\_tokenize} method to count the number of tokens in the document. Tokens from this tokenization do not necessarily match tokens from ChatGPT, as ChatGPT's tokenisation is similar to that in BERT (BERT tokenization matches candidate tokens against a controlled vocabulary and dividing tokens into matching subtokens out of vocabulary.) Through experimentation, we identified that a limit of 2,200 NLTK tokens from the document was about the maximum we could use to concatenate with the remainder of the prompt and issue to ChatGPT without encountering problems with the input size limit.

\begin{figure}

  \begin{verbatim}[question_text]

A web search for this question has returned the 
following evidence, which I provide to you in quotes:

"[passage_text]"

You MUST answer to my question with one of the following 
options ONLY: <Yes>, <No>. Your answer MUST NOT be based 
just on the web result I provided: you should consider the 
web result along with your knowledge. Please also provide 
an explanation for your answer.\end{verbatim}
  \caption{ChatGPT Prompt used to determine what impact a supportive or contrary passage has on answer correctness.}
  \label{fig:rq2_prompt}
\end{figure}

\subsection{Results}

\begin{figure}     
     \begin{subfigure}{0.66\columnwidth}
	  \includegraphics[width=1\linewidth]{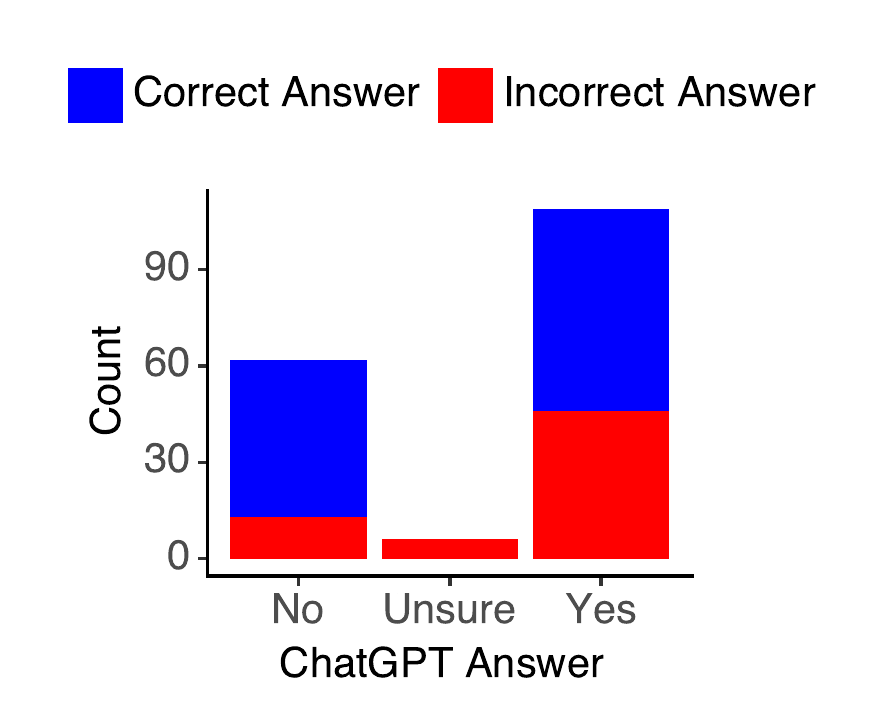}
     \end{subfigure}
     \begin{subfigure}{0.3\columnwidth}
         \centering
			\begin{tabular}{lr}
			\toprule
			  Correct   & 112        \\
 Unsure    &   6        \\
 Incorrect &  59        \\
 Total     & 177        \\
 Accuracy  &   0.63 \\
			\bottomrule
			\end{tabular}	         
      \end{subfigure}
  \caption{Effectiveness of ChatGPT when prompting with either a supporting or contrary evidence passage from TREC Misinformation qrels. }
  \label{fig:rq2_results}
\end{figure}

The effectiveness of ChatGPT with evidence-biased prompting is shown in Figure~\ref{fig:rq2_results}. Accuracy was less than the simpler, question only prompt (as shown previously in Figure~\ref{fig:rq1_results}).

\begin{figure}[t]
	\includegraphics[width=.8\columnwidth]{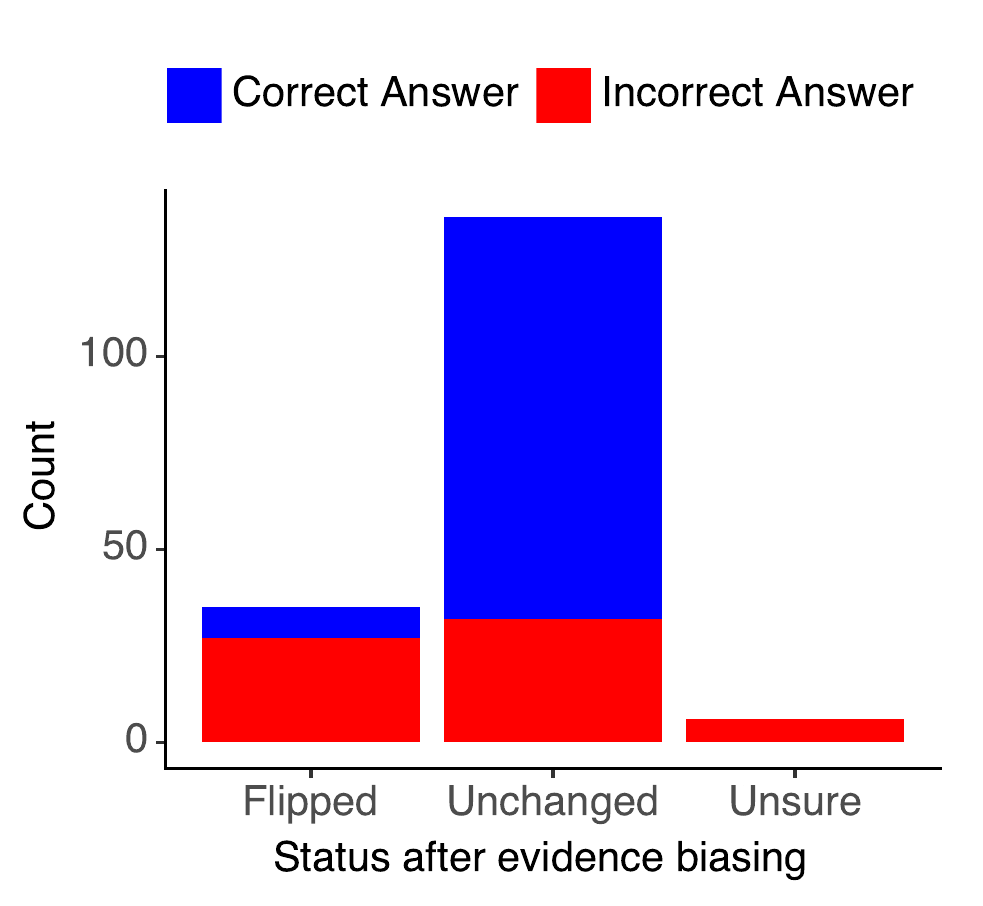}
	\caption{}
	\label{fig:rq2_flipped}
\end{figure}

\begin{figure}
  \includegraphics[width=1\columnwidth]{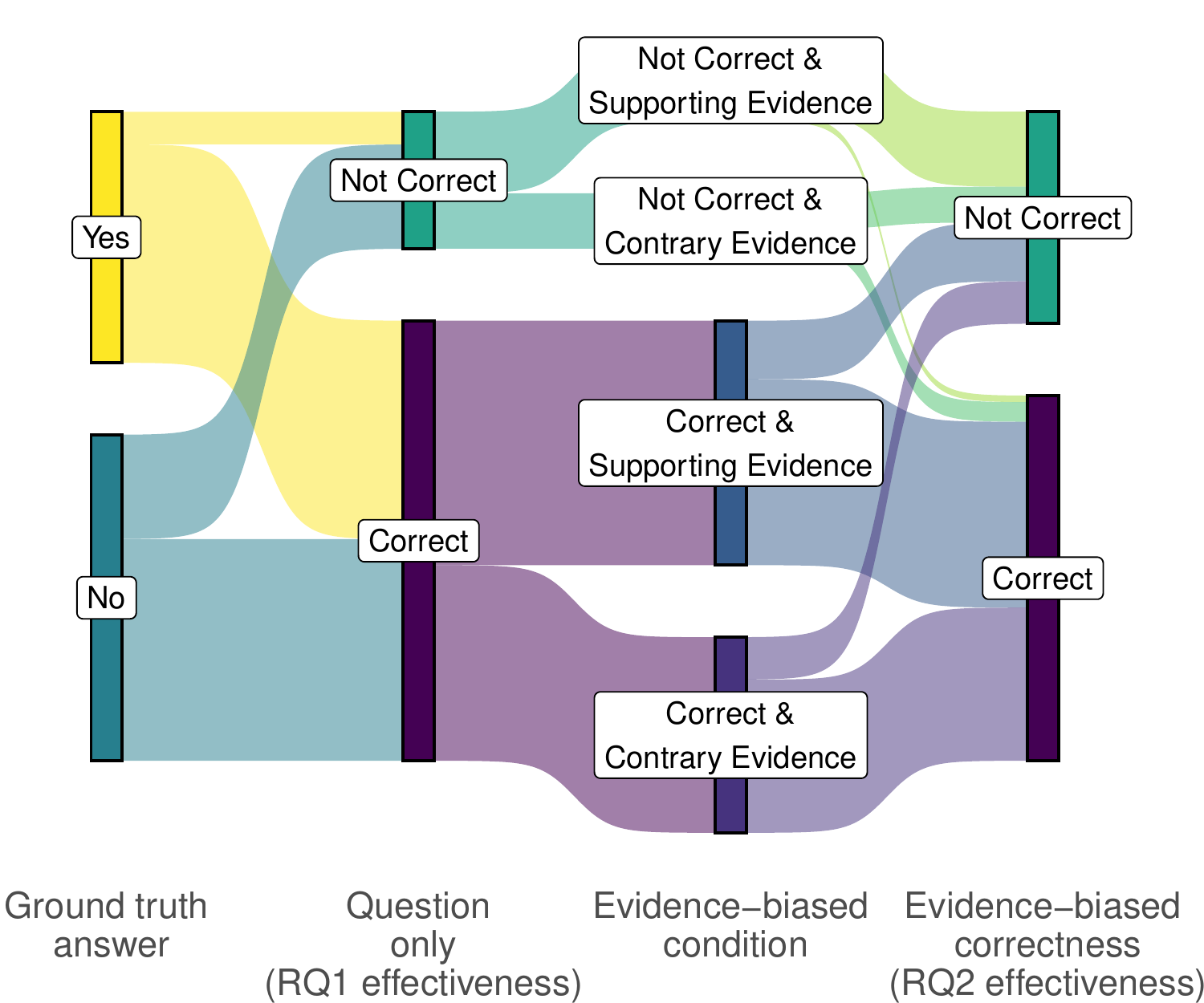}
  \caption{Sankey diagram showing the overall breakdown of all results. From the left, topics are divided by ground truth answer (Yes or No); next topics are divided according to RQ1 question prompting (Correct or Not Correct); next the prompt is evidence biased (Supporting and Contrary); finally, post evidence-biased breakdown is shown.}
  	\label{fig:rq2_sankey}
\end{figure}

Figure~\ref{fig:rq2_flipped} shows a breakdown of effectiveness after evidence-biasing. Specifically, this indicates how the answer changed compared to RQ1's question-only condition: "Flipped" indicates that ChatGPT's answered opposite after evidence-biasing; "Unchanged" means the answer matched RQ1's question-only condition; "Unsure" indicates where ChatGPT did not provide a Yes/No answer. We observe that when ChatGPT changes it's answer (i.e., Flipped), it generally gets the answer wrong.

Figure~\ref{fig:rq2_sankey} provides a detailed analysis of the answer behaviour of ChatGPT with respect to the two conditions: question-only and evidence-biased. This analysis provides a notable insight. Consider the incorrect answers provided in the evidence-biased condition (top-right condition in the Sankei diagram). First we note that there are little changes for the evidence to correct an answer that the model in the question-only condition got wrong. Then, we note that about half of the errors occur regardless of any evidence being provided -- that is, they occurred also in the question-only condition (the two top green lines). Thus, in these cases, the evidence had no effect in changing the model's response, regardless of whether the stance of the evidence matched the ground truth correctness. The other half of the errors instead are cases in which the question-only setting provided the correct answer, and instead the evidence-based prompt steered the model to provide an incorrect response. This occurs more often when supporting evidence is provided, rather than contrary evidence, but this difference is minimal.

%% file: Sections/6_conclusions.tex
\section{Conclusions}
In this paper we have examined the correctness of ChatGPT, a rapidly emerging large pre-trained generative language model, when answering complex health information questions regarding the effectiveness of a treatment for a condition. We did this in two settings: when only the question is presented to ChatGPT (question-only) and when the question is presented along with evidence (evidence-biased), i.e. a web search results retrieved when searching for the question. Importantly, we controlled whether the evidence is in favour or against the treatment. This in turn allowed us to understand the effect of providing evidence in the prompt and the impact that prompt knowledge can have on model knowledge and consequently on the generated answer. We found that ChatGPT answers correctly $80$\% of the questions if relying solely on model knowledge. On the other hand, the evidence presented in the prompt can heavily influence the answer -- and more importantly, it does affect the correctness of the answer, reducing ChatGPT's accuracy in our task to only $63$\%. Often, in fact, if evidence is provided, the model tends to agree with the stance of the evidence even if the model would produce an answer with the contrary stance if the evidence was not provided in input.

Our study has a number of limitations we plan to address in future work. 
In ChatGPT, like in other GPLMs, answer generation is stochastic. However, we did not study the variability of answer generation (and the associated effectiveness) across multiple runs of the same question. Similarly, we did not analyse what are the characteristics of the evidence we insert in prompts that  trigger divergent answers, despite having identical stance, nor we studied the effect of different prompt formats (including the extraction of key passages from the evidence documents used) -- aspects that are know to lead to variability in effectiveness in prompt learning~\cite{zhao2021calibrate}.
 
 A key feature of ChatGPT is its interaction abilities: it can hold a multi-turn conversation. In our experiments we discounted the multi-turn setting and only considered single-turn answers. The ability to hold multi-turn conversations would allow to e.g., provide multiple evidences, demand for a binary decision instead of and unsure position, and clarify aspects of the answer that may be unclear. 
 
 Finally, when producing an answer, we instructed ChatGPT to also explain the reason for providing the specific advice. These explanations often included claims about research studies and medical practices, which we did not validate whether their hold or are hallucinations of the model. In addition, we did not ask to attribute such claims to sources~\cite{bohnet2022attributed,menick2022teaching,rashkin2021measuring}: correct attribution appears to be a critical requirement to improve the quality, interpretability and  trustworthiness of these methods.

%% file: sigir2023-chatGPT-eliciting-model-knowledge-vs-prompt-knowledge.bbl

\begin{thebibliography}{13}


\ifx \showCODEN    \undefined \def \showCODEN     #1{\unskip}     \fi
\ifx \showDOI      \undefined \def \showDOI       #1{#1}\fi
\ifx \showISBNx    \undefined \def \showISBNx     #1{\unskip}     \fi
\ifx \showISBNxiii \undefined \def \showISBNxiii  #1{\unskip}     \fi
\ifx \showISSN     \undefined \def \showISSN      #1{\unskip}     \fi
\ifx \showLCCN     \undefined \def \showLCCN      #1{\unskip}     \fi
\ifx \shownote     \undefined \def \shownote      #1{#1}          \fi
\ifx \showarticletitle \undefined \def \showarticletitle #1{#1}   \fi
\ifx \showURL      \undefined \def \showURL       {\relax}        \fi
\providecommand\bibfield[2]{#2}
\providecommand\bibinfo[2]{#2}
\providecommand\natexlab[1]{#1}
\providecommand\showeprint[2][]{arXiv:#2}

\bibitem[\protect\citeauthoryear{Bang, Cahyawijaya, Lee, Dai, Su, Wilie,
  Lovenia, Ji, Yu, Chung, et~al\mbox{.}}{Bang et~al\mbox{.}}{2023}]%
        {bang2023multitask}
\bibfield{author}{\bibinfo{person}{Yejin Bang}, \bibinfo{person}{Samuel
  Cahyawijaya}, \bibinfo{person}{Nayeon Lee}, \bibinfo{person}{Wenliang Dai},
  \bibinfo{person}{Dan Su}, \bibinfo{person}{Bryan Wilie},
  \bibinfo{person}{Holy Lovenia}, \bibinfo{person}{Ziwei Ji},
  \bibinfo{person}{Tiezheng Yu}, \bibinfo{person}{Willy Chung},
  {et~al\mbox{.}}} \bibinfo{year}{2023}\natexlab{}.
\newblock \showarticletitle{A Multitask, Multilingual, Multimodal Evaluation of
  ChatGPT on Reasoning, Hallucination, and Interactivity}.
\newblock \bibinfo{journal}{\emph{arXiv preprint arXiv:2302.04023}}
  (\bibinfo{year}{2023}).
\newblock


\bibitem[\protect\citeauthoryear{Benoit}{Benoit}{2023}]%
        {benoit2023chatgpt}
\bibfield{author}{\bibinfo{person}{James~RA Benoit}.}
  \bibinfo{year}{2023}\natexlab{}.
\newblock \showarticletitle{ChatGPT for Clinical Vignette Generation, Revision,
  and Evaluation}.
\newblock \bibinfo{journal}{\emph{medRxiv}} (\bibinfo{year}{2023}),
  \bibinfo{pages}{2023--02}.
\newblock


\bibitem[\protect\citeauthoryear{Bohnet, Tran, Verga, Aharoni, Andor, Soares,
  Eisenstein, Ganchev, Herzig, Hui, et~al\mbox{.}}{Bohnet
  et~al\mbox{.}}{2022}]%
        {bohnet2022attributed}
\bibfield{author}{\bibinfo{person}{Bernd Bohnet}, \bibinfo{person}{Vinh~Q
  Tran}, \bibinfo{person}{Pat Verga}, \bibinfo{person}{Roee Aharoni},
  \bibinfo{person}{Daniel Andor}, \bibinfo{person}{Livio~Baldini Soares},
  \bibinfo{person}{Jacob Eisenstein}, \bibinfo{person}{Kuzman Ganchev},
  \bibinfo{person}{Jonathan Herzig}, \bibinfo{person}{Kai Hui},
  {et~al\mbox{.}}} \bibinfo{year}{2022}\natexlab{}.
\newblock \showarticletitle{Attributed Question Answering: Evaluation and
  Modeling for Attributed Large Language Models}.
\newblock \bibinfo{journal}{\emph{arXiv preprint arXiv:2212.08037}}
  (\bibinfo{year}{2022}).
\newblock


\bibitem[\protect\citeauthoryear{Chen, Fisch, Weston, and Bordes}{Chen
  et~al\mbox{.}}{2017}]%
        {chen2017reading}
\bibfield{author}{\bibinfo{person}{Danqi Chen}, \bibinfo{person}{Adam Fisch},
  \bibinfo{person}{Jason Weston}, {and} \bibinfo{person}{Antoine Bordes}.}
  \bibinfo{year}{2017}\natexlab{}.
\newblock \showarticletitle{Reading Wikipedia to Answer Open-Domain Questions}.
  In \bibinfo{booktitle}{\emph{Proceedings of the 55th Annual Meeting of the
  Association for Computational Linguistics (Volume 1: Long Papers)}}.
  \bibinfo{pages}{1870--1879}.
\newblock


\bibitem[\protect\citeauthoryear{De~Angelis, Baglivo, Arzilli, Privitera,
  Ferragina, Tozzi, and Rizzo}{De~Angelis et~al\mbox{.}}{2023}]%
        {de2023chatgpt}
\bibfield{author}{\bibinfo{person}{Luigi De~Angelis},
  \bibinfo{person}{Francesco Baglivo}, \bibinfo{person}{Guglielmo Arzilli},
  \bibinfo{person}{Gaetano~Pierpaolo Privitera}, \bibinfo{person}{Paolo
  Ferragina}, \bibinfo{person}{Alberto~Eugenio Tozzi}, {and}
  \bibinfo{person}{Caterina Rizzo}.} \bibinfo{year}{2023}\natexlab{}.
\newblock \showarticletitle{ChatGPT and the Rise of Large Language Models: The
  New AI-Driven Infodemic Threat in Public Health}.
\newblock \bibinfo{journal}{\emph{Available at SSRN 4352931}}
  (\bibinfo{year}{2023}).
\newblock


\bibitem[\protect\citeauthoryear{Gilson, Safranek, Huang, Socrates, Chi,
  Taylor, and Chartash}{Gilson et~al\mbox{.}}{2022}]%
        {gilson2022well}
\bibfield{author}{\bibinfo{person}{Aidan Gilson}, \bibinfo{person}{Conrad
  Safranek}, \bibinfo{person}{Thomas Huang}, \bibinfo{person}{Vimig Socrates},
  \bibinfo{person}{Ling Chi}, \bibinfo{person}{Richard~Andrew Taylor}, {and}
  \bibinfo{person}{David Chartash}.} \bibinfo{year}{2022}\natexlab{}.
\newblock \showarticletitle{How Well Does ChatGPT Do When Taking the Medical
  Licensing Exams? The Implications of Large Language Models for Medical
  Education and Knowledge Assessment}.
\newblock \bibinfo{journal}{\emph{medRxiv}} (\bibinfo{year}{2022}),
  \bibinfo{pages}{2022--12}.
\newblock


\bibitem[\protect\citeauthoryear{Ji, Lee, Frieske, Yu, Su, Xu, Ishii, Bang,
  Madotto, and Fung}{Ji et~al\mbox{.}}{2022}]%
        {ji2022survey}
\bibfield{author}{\bibinfo{person}{Ziwei Ji}, \bibinfo{person}{Nayeon Lee},
  \bibinfo{person}{Rita Frieske}, \bibinfo{person}{Tiezheng Yu},
  \bibinfo{person}{Dan Su}, \bibinfo{person}{Yan Xu}, \bibinfo{person}{Etsuko
  Ishii}, \bibinfo{person}{Yejin Bang}, \bibinfo{person}{Andrea Madotto}, {and}
  \bibinfo{person}{Pascale Fung}.} \bibinfo{year}{2022}\natexlab{}.
\newblock \showarticletitle{Survey of hallucination in natural language
  generation}.
\newblock \bibinfo{journal}{\emph{Comput. Surveys}} (\bibinfo{year}{2022}).
\newblock


\bibitem[\protect\citeauthoryear{Karpukhin, Oguz, Min, Lewis, Wu, Edunov, Chen,
  and Yih}{Karpukhin et~al\mbox{.}}{2020}]%
        {karpukhin2020dense}
\bibfield{author}{\bibinfo{person}{Vladimir Karpukhin}, \bibinfo{person}{Barlas
  Oguz}, \bibinfo{person}{Sewon Min}, \bibinfo{person}{Patrick Lewis},
  \bibinfo{person}{Ledell Wu}, \bibinfo{person}{Sergey Edunov},
  \bibinfo{person}{Danqi Chen}, {and} \bibinfo{person}{Wen-tau Yih}.}
  \bibinfo{year}{2020}\natexlab{}.
\newblock \showarticletitle{Dense Passage Retrieval for Open-Domain Question
  Answering}. In \bibinfo{booktitle}{\emph{Proceedings of the 2020 Conference
  on Empirical Methods in Natural Language Processing (EMNLP)}}.
  \bibinfo{pages}{6769--6781}.
\newblock


\bibitem[\protect\citeauthoryear{Liu, Yuan, Fu, Jiang, Hayashi, and Neubig}{Liu
  et~al\mbox{.}}{2023}]%
        {liu2023pre}
\bibfield{author}{\bibinfo{person}{Pengfei Liu}, \bibinfo{person}{Weizhe Yuan},
  \bibinfo{person}{Jinlan Fu}, \bibinfo{person}{Zhengbao Jiang},
  \bibinfo{person}{Hiroaki Hayashi}, {and} \bibinfo{person}{Graham Neubig}.}
  \bibinfo{year}{2023}\natexlab{}.
\newblock \showarticletitle{Pre-train, prompt, and predict: A systematic survey
  of prompting methods in natural language processing}.
\newblock \bibinfo{journal}{\emph{Comput. Surveys}} \bibinfo{volume}{55},
  \bibinfo{number}{9} (\bibinfo{year}{2023}), \bibinfo{pages}{1--35}.
\newblock


\bibitem[\protect\citeauthoryear{Menick, Trebacz, Mikulik, Aslanides, Song,
  Chadwick, Glaese, Young, Campbell-Gillingham, Irving, et~al\mbox{.}}{Menick
  et~al\mbox{.}}{2022}]%
        {menick2022teaching}
\bibfield{author}{\bibinfo{person}{Jacob Menick}, \bibinfo{person}{Maja
  Trebacz}, \bibinfo{person}{Vladimir Mikulik}, \bibinfo{person}{John
  Aslanides}, \bibinfo{person}{Francis Song}, \bibinfo{person}{Martin
  Chadwick}, \bibinfo{person}{Mia Glaese}, \bibinfo{person}{Susannah Young},
  \bibinfo{person}{Lucy Campbell-Gillingham}, \bibinfo{person}{Geoffrey
  Irving}, {et~al\mbox{.}}} \bibinfo{year}{2022}\natexlab{}.
\newblock \showarticletitle{Teaching language models to support answers with
  verified quotes}.
\newblock \bibinfo{journal}{\emph{arXiv preprint arXiv:2203.11147}}
  (\bibinfo{year}{2022}).
\newblock


\bibitem[\protect\citeauthoryear{Nov, Singh, and Mann}{Nov
  et~al\mbox{.}}{2023}]%
        {nov2023putting}
\bibfield{author}{\bibinfo{person}{Oded Nov}, \bibinfo{person}{Nina Singh},
  {and} \bibinfo{person}{Devin~M Mann}.} \bibinfo{year}{2023}\natexlab{}.
\newblock \showarticletitle{Putting ChatGPT's Medical Advice to the (Turing)
  Test}.
\newblock \bibinfo{journal}{\emph{medRxiv}} (\bibinfo{year}{2023}),
  \bibinfo{pages}{2023--01}.
\newblock


\bibitem[\protect\citeauthoryear{Rashkin, Nikolaev, Lamm, Collins, Das, Petrov,
  Tomar, Turc, and Reitter}{Rashkin et~al\mbox{.}}{2021}]%
        {rashkin2021measuring}
\bibfield{author}{\bibinfo{person}{Hannah Rashkin}, \bibinfo{person}{Vitaly
  Nikolaev}, \bibinfo{person}{Matthew Lamm}, \bibinfo{person}{Michael Collins},
  \bibinfo{person}{Dipanjan Das}, \bibinfo{person}{Slav Petrov},
  \bibinfo{person}{Gaurav~Singh Tomar}, \bibinfo{person}{Iulia Turc}, {and}
  \bibinfo{person}{David Reitter}.} \bibinfo{year}{2021}\natexlab{}.
\newblock \showarticletitle{Measuring attribution in natural language
  generation models}.
\newblock \bibinfo{journal}{\emph{arXiv preprint arXiv:2112.12870}}
  (\bibinfo{year}{2021}).
\newblock


\bibitem[\protect\citeauthoryear{Zhao, Wallace, Feng, Klein, and Singh}{Zhao
  et~al\mbox{.}}{2021}]%
        {zhao2021calibrate}
\bibfield{author}{\bibinfo{person}{Zihao Zhao}, \bibinfo{person}{Eric Wallace},
  \bibinfo{person}{Shi Feng}, \bibinfo{person}{Dan Klein}, {and}
  \bibinfo{person}{Sameer Singh}.} \bibinfo{year}{2021}\natexlab{}.
\newblock \showarticletitle{Calibrate before use: Improving few-shot
  performance of language models}. In \bibinfo{booktitle}{\emph{International
  Conference on Machine Learning}}. PMLR, \bibinfo{pages}{12697--12706}.
\newblock


\end{thebibliography}
